\documentclass{article}

\usepackage[preprint]{neurips_2025}

\usepackage[utf8]{inputenc} 
\usepackage[T1]{fontenc}    
\usepackage{hyperref}       
\usepackage{url}            
\usepackage{booktabs}       
\usepackage{amsfonts}       
\usepackage{nicefrac}       
\usepackage{microtype}      
\usepackage{xcolor}         

\usepackage{subcaption}
\usepackage{graphicx}
\usepackage{amsmath}
\usepackage{amssymb}
\usepackage{adjustbox}
\usepackage{makecell}
\usepackage{multirow}
\usepackage{bbding}
\usepackage{color, colortbl}
\usepackage{enumitem}

\usepackage{amsthm}

\usepackage{circledsteps}

\usepackage{pifont}

\usepackage{algorithm}
\usepackage{algorithmicx}
\usepackage{algpseudocode}

\title{Quantization Meets dLLMs: A Systematic Study of Post-training Quantization for Diffusion LLMs}

\author{Haokun Lin$^{*\ 1,3}$, Haobo Xu$^{*\ 2}$, 
Yichen Wu$^{3,4}$, 
Ziyu Guo$^{5}$, Renrui Zhang$^5$
\vspace{0.1cm},\\
\textbf{Zhichao Lu$^3$, Ying Wei$^6$, Qingfu Zhang$^3$,
Zhenan Sun$^{1}$ }\vspace{0.2cm}\\
  $^*$Equal Contribution\vspace{0.2cm} \\
    $^1$ NLPR \& MAIS, Institute of Automation, CAS \quad
    $^2$ Tsinghua University \\
    $^3$ City University of Hong Kong \quad
    $^4$ Harvard University \\
    $^5$ The Chinese University of Hong Kong \quad
    $^6$ Zhejiang University \\
    \vspace{-0.3cm}
}

\begin{document}

\maketitle

\begin{abstract}
    Recent advances in diffusion large language models (dLLMs) have introduced a promising alternative to autoregressive (AR) LLMs for natural language generation tasks, leveraging full attention and denoising-based decoding strategies. However, the deployment of these models on edge devices remains challenging due to their massive parameter scale and high resource demands. While post-training quantization (PTQ) has emerged as a widely adopted technique for compressing AR LLMs, its applicability to dLLMs remains largely unexplored. In this work, we present the first systematic study on quantizing diffusion-based language models. We begin by identifying the presence of activation outliers, characterized by abnormally large activation values that dominate the dynamic range. These outliers pose a key challenge to low-bit quantization, as they make it difficult to preserve precision for the majority of values.
    More importantly, we implement state-of-the-art PTQ methods and conduct a comprehensive evaluation across multiple task types and model variants. Our analysis is structured along four key dimensions: bit-width, quantization method, task category, and model type. Through this multi-perspective evaluation, we offer practical insights into the quantization behavior of dLLMs under different configurations. We hope our findings provide a foundation for future research in efficient dLLM deployment. 
    Our code is publicly available at \url{https://github.com/FelixMessi/QDLM}.
\end{abstract}
\section{Introduction}

Large language models (LLMs) have achieved remarkable success in a wide range of text generation tasks, with auto-regressive architectures—such as GPT~\citep{brown2020gpt,brown2020gpt3,achiam2023gpt4}, LLaMA~\citep{touvron2023llama,touvron2023llama2,dubey2024llama3}, and the Qwen~\citep{bai2023qwen,qwen2025qwen25technicalreport,yang2025qwen3} series—dominating recent advances in both research and application.
Recently, diffusion-based large language models (dLLMs) have emerged as a promising alternative for natural language generation~\citep{nie2025llada,zhu2025llada1.5,dream2025,gong2024diffullama,song2025seed-diffusion}. 
By leveraging bidirectional context encoding and iterative denoising, dLLMs offer finer-grained control over the generation process compared to traditional auto-regressive approaches.
Despite their potential, the efficient deployment of dLLMs remains challenging, as the increased number of model parameters often leads to significantly higher memory usage and computational cost~\citep{li2025survey_dllm,yu2025discretediffusionlargelanguage}.

Current efforts toward optimizing dLLM inference have primarily focused on designing specialized key-value (KV) cache mechanisms ~\citep{wu2025fastdllm,ma2025dkvcache,liu2025dllmcache,wang2025df-dllm}. However, quantization ~\citep{li2024eval_quant,liu2025quant_reasoning,wei2025roste,ye2025dbellquant}, a well-established yet orthogonal technique for compressing and accelerating neural networks, has been largely underexplored in the context of dLLMs.
In the domain of auto-regressive LLMs, post-training quantization(PTQ)~\citep{chee2024quip,ashkboos2023quik,tseng2024quip_sharp,zhao2023atom} has been widely adopted to reduce the memory footprint of weights and activations, and to enable faster inference through kernel-level optimization. 
Yet, how well existing PTQ techniques generalize to diffusion LLMs remains an open and intriguing question.

\begin{figure*}[t]
    \centering
    \includegraphics[width=0.98\linewidth]{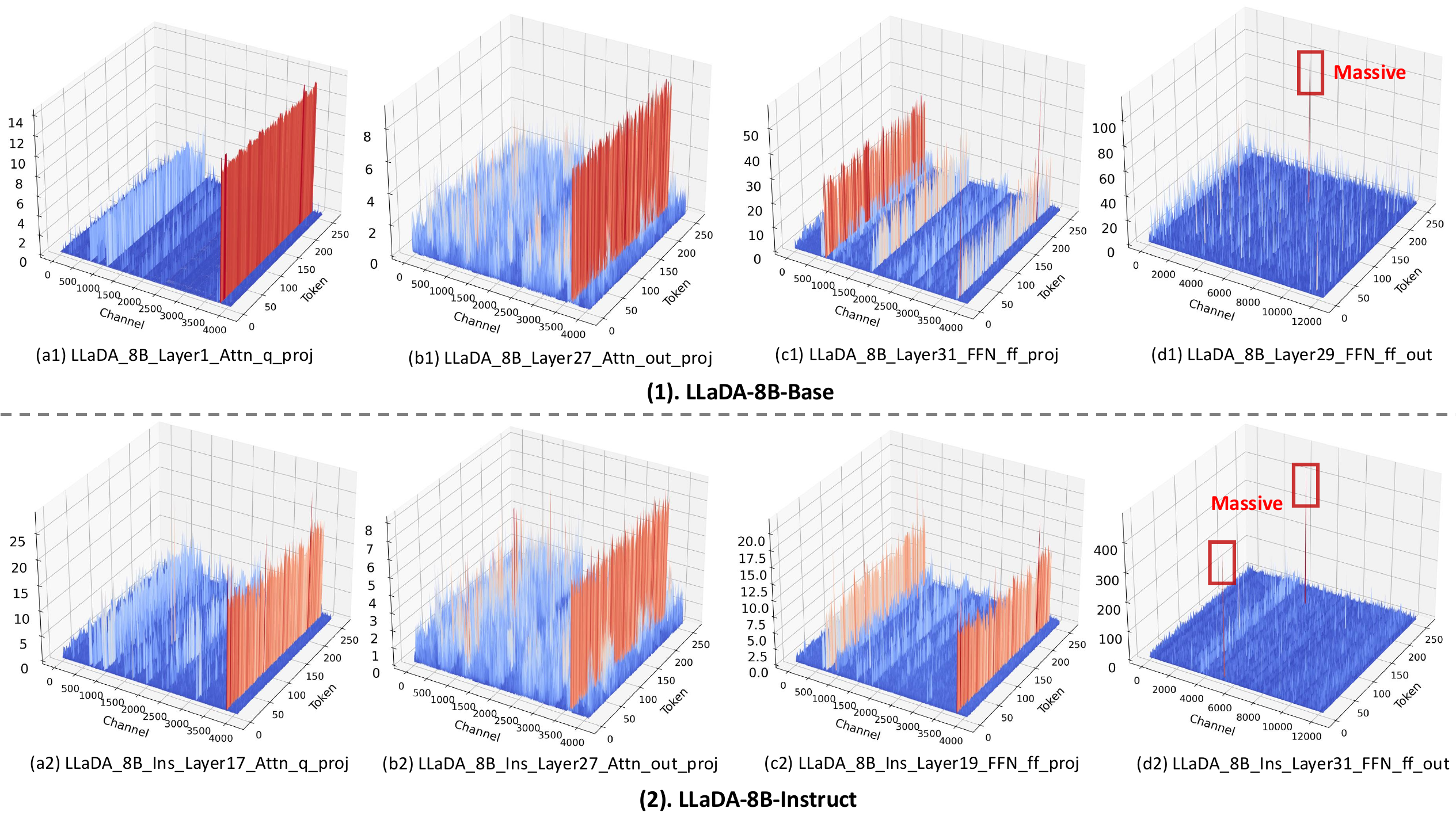}
    \caption{
Visualizations of activation outliers in LLaDA-8B-Base (1) and LLaDA-8B-Instruct (2). Outliers are observed at the inputs of various linear layers and can be classified as Normal Outliers (a(1)–c(1)/a(2)–c(2)), with relatively large magnitudes across tokens, and Massive Outliers (d(1), d(2)), with extremely large values on a few tokens. Notably, these massive outliers are identified at the second linear layer of the feed-forward network (FFN) module.
    }
    \label{fig:outlier_vis}    
\end{figure*}

\begin{figure*}[t]
    \centering
    \includegraphics[width=1.\linewidth]{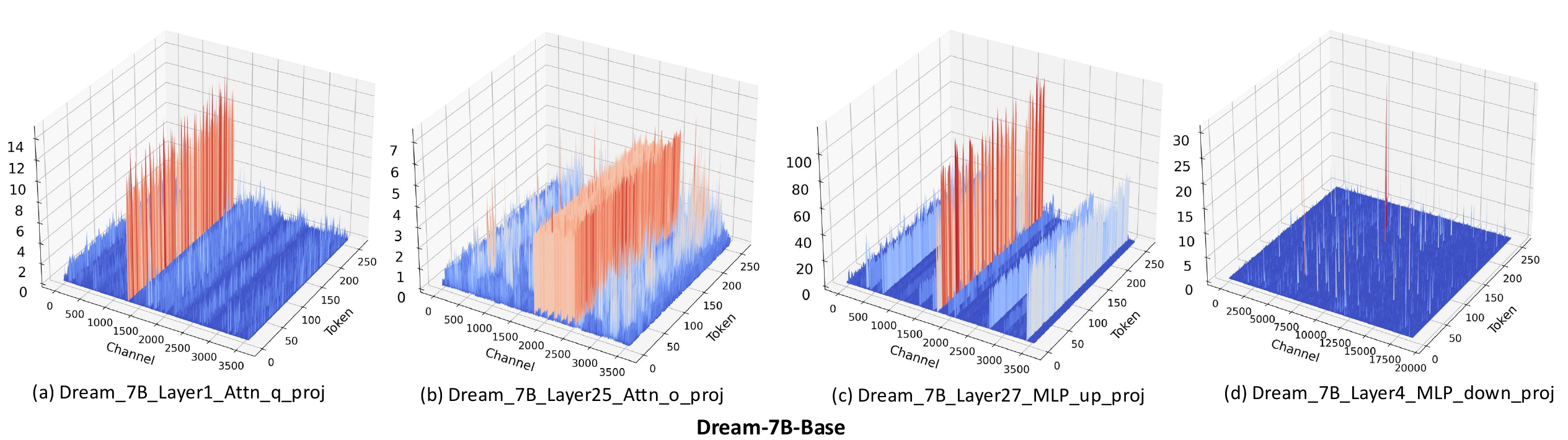}
    \caption{
Visualizations of activation outliers in Dream-7B-Base.
We observe relatively large normal outliers in the input to the FFN up-projection layer (c), while the massive outliers (d) exhibit smaller peak values compared to those in LLaDA models (Figure~\ref{fig:outlier_vis}).
    }
    \label{fig:outlier_vis_dream}    
\end{figure*}

In this paper, we present a comprehensive study on the quantization of diffusion-based large language models (dLLMs).
First, we identify that dLLMs exhibit clear activation outliers—i.e., unusually large activation values—which are known to be a key challenge for low-bit quantization~\citep{dettmers2022llm_int8,xiao2023smoothquant,sun2024massive}.
Specifically, as shown in Figure~\ref{fig:outlier_vis} and~\ref{fig:outlier_vis_dream},
we observe such outliers across multiple layers and input activations in LLaDA-Base, LLaDA-Instruct~\citep{nie2025llada}, and Dream~\citep{dream2025} models, suggesting that this is a common phenomenon across different dLLMs.
Second, we implement state-of-the-art weight-only~\citep{lin2023awq,frantar2022gptq} and weight-activation quantization~\citep{xiao2023smoothquant,ashkboos2024quarot,lin2024duquant} methods on representative diffusion models and conduct a detailed analysis from the following perspectives:

\begin{itemize}[leftmargin=10pt]
    \item \textbf{Bit-width effects}: We find that 4-bit is the most effective configuration for weight-only quantization, while 8-bit is recommended for weight-activation quantization as a near-lossless setting.
    \item \textbf{Quantization methods}: Through extensive evaluation, we observe that GPTQ consistently outperforms AWQ across most tasks. For weight-activation quantization, rotation-based methods such as DuQuant and QuaRot demonstrate clear advantages over SmoothQuant.
    \item \textbf{Task type sensitivity}: While most PTQ methods perform competitively on general QA benchmarks, we observe notable degradation on more complex tasks such as math reasoning and code generation.
    \item \textbf{Model type robustness}: Our results show that the instruction-tuned LLaDA model exhibits greater robustness to quantization compared to the base counterpart.
\end{itemize}

To the best of our knowledge, this is the first systematic evaluation of post-training quantization on diffusion LLMs. We hope our findings provide valuable guidance for the community and inspire further research toward efficient and deployable dLLMs.
\section{Related Work}
\label{sec:related}

\subsection{Diffusion Language Model}

With the fast development of deep learning and pre-trained models~\citep{zeng2023generative,zeng2023parrot,zeng2024hierarchical,zeng2024interformer,zeng2025pave,yan2021bright,yan2022dissecting,yan2021dynamic,yan2023trainable,yanred,yan2023reconciling,yan2024pacer,yan2024thegcn,yan2024topological,lin2025toklip}, 
Diffusion models have achieved remarkable success in image, video, and audio generation by learning to reverse a forward noise process~\citep{jiangimage,zhao2024vidit}. However, applying diffusion to language generation presents unique challenges due to the discrete nature of textual data. To address this, DiffusionBERT~\citep{he2022diffusionbert} leverages a BERT~\citep{devlin2019bert} architecture to model the reverse dynamics of a discrete diffusion process with an absorbing state, as proposed in D3PM~\citep{austin2021d3pm}.

More recently, Masked Diffusion Models (MDMs)~\citep{lou2023sedd,ou2024your,shi2024simplified} have drawn increasing attention by adopting a forward process that progressively replaces input tokens with a designated [MASK] token. 
This year, efforts have been made to scale up MDMs to the billion-parameter regime. Representative examples include LLaDA-8B~\citep{nie2025llada}, which utilizes a bidirectional Transformer as the mask denoiser and achieves performance comparable to LLaMA~\citep{dubey2024llama3}, and Dream~\citep{dream2025}, which is initialized from a pre-trained autoregressive model and delivers competitive generation capabilities. These advancements indicate that diffusion-based approaches offer a viable alternative paradigm for language modeling.

Despite these encouraging results, the deployment of diffusion large language models (dLLMs)~\citep{gong2024diffullama,yang2025mmada} remains constrained by the computational demands of Transformer-based architectures, which involve hundreds of millions of parameters. To address this, we explore the potential of extending established post-training quantization techniques from conventional LLMs to the dLLM models, aiming to reduce memory footprint and accelerate inference while preserving generation quality.
Notably, some recent works~\citep{wu2025fastdllm,liu2025dllmcache,ma2025dkvcache} propose caching strategies to accelerate the inference of dLLMs. Our work is orthogonal to these efforts and can be seamlessly integrated by quantizing dLLM caches.

\subsection{Network Quantization}

Compared to pruning and distillation~\citep{lin2024mope,zhang2024plug,xing2025efficientllm,ye2025voco,ye2025atp},
quantization has been extensively studied as an effective technique to compress neural networks by using low-bit representations for high-precision tensors~\citep{zhao2024mixdq,xu2024slog,yang2025lrq,huang2025quaff}
Existing methods are typically categorized into two groups: post-training quantization (PTQ) and quantization-aware training (QAT). PTQ~\citep{wu2024ptq4dit,yang2024dopq,xu2025dllmquant} applies quantization after model training, while QAT~\citep{tao2022compression_gpt,chen2024efficientqat,chen2025scalinglaw} incorporates quantization effects during training.
Due to the high computational cost of training large language models (LLMs), PTQ has become increasingly popular for its efficiency and ability to preserve model performance without retraining~\citep{liu2024intactkv,dong2024stbllm}.
In this work, we follow this paradigm and focus on applying PTQ to dLLMs.

\textbf{Weight-only quantization} compresses the model by quantizing weight matrices, effectively reducing model size and memory access during inference. For example, GPTQ~\citep{frantar2022gptq} extends the Optimal Brain Quantization~\citep{frantar2022obq} algorithm to LLMs, AWQ~\citep{lin2023awq} introduces a reparameterization strategy to alleviate the difficulty of weight quantization, and SqueezeLLM~\citep{kim2023squeezellm} employs non-uniform quantization to improve compression quality.

\textbf{Weight-activation quantization} quantizes both the model weights and input activations, enabling further inference acceleration by leveraging integer matrix multiplication kernels.
SmoothQuant~\citep{xiao2023smoothquant} proposes to shift the quantization difficulty from activations to weights via scaling. 
OmniQuant~\citep{shao2023omniquant} jointly optimizes clipping thresholds and scaling factors for improved quantization fidelity.
More recently, rotation-based methods~\citep{lin2024qserve,hu2025ostquant} have demonstrated superior performance: QuaRot~\citep{ashkboos2024quarot} introduces Hadamard-based rotation to smooth the weight-activation landscape, while DuQuant~\citep{lin2024duquant} leverages outlier-aware rotation matrices and channel permutation to better align the activation distribution with quantization-friendly structures.

In this work,
we provide a comprehensive evaluation of state-of-the-art LLM-oriented PTQ methods applied to diffusion-based language models. All methods are re-implemented on dLLMs, and we present in-depth analyses and insights into their quantization performance.

\section{Preliminary and Observation}

\subsection{Masked Diffusion Model}
Masked diffusion model is a variant of diffusion-based generative models that incorporates a binary mask into the denoising process. Instead of reconstructing the entire input, the model focuses on predicting the corrupted or missing regions while preserving the observed parts. Specifically, given an input $x$ and a mask $m$, the forward process adds Gaussian noise to the unmasked regions, producing a noised sample $x_t$ at step $t$. The reverse process is then parameterized by a neural network $\epsilon_\theta$, which estimates the noise conditioned on both the timestep and the mask. The training objective is,
$$
\mathcal{L}_{\text{MDM}} = 
\mathbb{E}_{x, m, \epsilon, t} 
\left[ \, \| \epsilon - \epsilon_\theta(x_t, m, t) \|^2 \, \right],
$$
where $\epsilon$ denotes the Gaussian noise, and $\epsilon_\theta$ learns to predict and remove it under the masking constraint.
\subsection{Quantization}

Quantization coverts the floating-point tensor $\mathbf{X}$ into a low-bit integer $\mathbf{X}_{q}$. Specifically, the $b$-bit uniform quantization can be represented as:

\begin{equation}
    \label{eq_quantization}
    \mathbf{X}_{q} = \text{clamp}\left(\left\lfloor \frac{\mathbf{X}}{s} \right\rceil \!\!+\! z, 0, 2^{b}-1 \right), 
    \textrm{where}~s=\frac{\max(\mathbf{X})-\min(\mathbf{X})}{2^b-1}, z = -\left\lfloor \frac{\min(\mathbf{X})}{s}\right\rceil. ~~~ 
\end{equation}   

The notation $\left\lfloor \cdot \right\rceil$ means the nearest rounding operation, $s$ is the quantization step size and $z$ denotes the zero point.

\subsection{Outliers in dLLMs}

Outliers, a prominent characteristic of large language models (LLMs), are primarily determined by relatively large activation values~\citep{dettmers2022llm_int8}. These outliers are typically categorized into two types: normal outliers and massive outliers~\citep{lin2024duquant}.
Normal outliers~\citep{xiao2023smoothquant} refer to activations across all tokens with relatively large magnitudes, and they are the more prevalent type.
Massive outliers~\citep{sun2024massive,liu2024intactkv}, on the other hand, exhibit significantly larger values at a limited set of tokens. These outliers present substantial challenges for LLM quantization.
Whether dLLMs contain these outliers remains an important yet under-explored question. In this work, we provide a detailed preliminary exploration and identify the presence of outliers in dLLMs.

We first identify the presence of activation outliers in diffusion-based language models. 
Specifically, we randomly sample a batch of calibration data from the WikiText-2 dataset~\citep{merity2016wiki} and use it as input for a single forward pass to visualize the activation distributions across different layers.
As shown in Figure~\ref{fig:outlier_vis}, we observe clear outliers in the input activations of both LLaDA-8B-base and LLaDA-8B-instruct. 
These outliers can be categorized into two types: Normal Outliers and Massive Outliers, consistent with the taxonomy observed in standard LLMs.
Interestingly, the Massive Outliers tend to occur in the second linear layer of the feed-forward network (FFN) modules, mirroring patterns reported in previous studies on conventional LLMs~\citep{an2025systematic}. 
However, compared to LLMs, the Normal Outliers in LLaDA exhibit slightly lower magnitudes, indicating a less extreme but still significant deviation.
Another key difference is that massive outliers in dLLMs appear across more tokens, rather than being restricted to only a few tokens as in LLMs.
This broader distribution increases the difficulty of weight-activation quantization, as it reduces the effectiveness of global clipping or scaling strategies.
This observation is corroborated by the near-zero performance of SmoothQuant under W4A4 settings (see Table~\ref{tab:wa-math-code}), suggesting that existing outlier-handling strategies may be insufficient for dLLMs in low-bit quantization regimes.
Furthermore, we also detect similar outlier patterns in the Dream-7B model, as visualized in Figure~\ref{fig:outlier_vis_dream}. 
This indicates that the existence of outliers is not specific to a particular model architecture, but rather a general phenomenon across diffusion-based language models.
These findings highlight the need for careful handling of outliers during the quantization process, especially when targeting both weights and activations under aggressive bit-width constraints.
\section{Quantizing Diffusion LLM}

In this section, we conduct experiments to address the overarching question: \textbf{How does quantization affect diffusion-based language models?}
To systematically explore this, we further investigate the following sub-questions:
\begin{itemize}
    \item \textbf{RQ1:} What are the preferred bit-widths for weight-only and weight-activation quantization?
    \item \textbf{RQ2:} What are the most effective quantization methods for dLLMs?
    \item \textbf{RQ3:} How do different task categories influence the performance of quantized dLLMs?
    \item \textbf{RQ4:} How does quantization affect different types of dLLMs?
\end{itemize}

\subsection{Experimental Setup}

\paragraph{Evaluated dLLMs and Quantization Baselines.}
We conduct comprehensive evaluations on three recent diffusion-based language models, LLaDA-8B-Base, LLaDA-8B-Instruct~\citep{nie2025llada} and Dream 7B-Base~\citep{dream2025}.
For \textit{weight-only quantization}, we adopt state-of-the-art baselines GPTQ~\citep{frantar2022gptq} and AWQ~\citep{lin2023awq}, which are widely used in LLM quantization. 
We utilize group-wise per-channel quantization and set the group size to 128.
For \textit{weight-activation quantization}, we evaluate SmoothQuant~\citep{xiao2023smoothquant} as well as recent rotation-based approaches, including QuaRot~\citep{ashkboos2024quarot} and DuQuant~\citep{lin2024duquant}. Following standard practice, we apply per-channel quantization to weights and per-token quantization to activations.
We select calibration data (128 samples) from WiKiText2~\citep{merity2016wiki} for baselines, except Pile~\citep{gao2020pile} for AWQ.
More details are illustrated in Appendix~\ref{appx:imple}.

\paragraph{Evaluation Benchmarks.}
We evaluate the performance of quantized dLLMs across three task categories, following the setup of LLaDA~\citep{nie2025llada}: 1). General knowledge tasks, including MMLU~\citep{hendrycks2020mmlu}, ARC-E, ARC-C~\citep{clark2018arc}, Hellaswag~\citep{zellers2019hellaswag}, WinoGrande~\citep{sakaguchi2021winogrande}, and PIQA~\citep{bisk2020piqa}; 2). Mathematical reasoning tasks, such as GSM8K~\citep{cobbe2021gsm8k} and Math~\citep{hendrycks2021math}; and 3).Code generation tasks, including HumanEval~\citep{chen2021humaneval} and MBPP~\citep{austin2021mbpp}.
These benchmarks collectively provide a comprehensive assessment of quantized dLLMs from multiple perspectives.

\paragraph{Evaluation Metrics.}
We report accuracy on widely used QA and math benchmarks, and adopt Pass@1 as the evaluation metric for code generation tasks.
Performance degradation relative to full-precision models is used as the primary metric for assessing different quantized dLLMs.
Following \citep{liu2025quant_reasoning}, we categorize the performance degradation compared to full-precision models into three levels: \textbf{negligible(<1\%)}, \textbf{moderate (1–4\%)}, and \textbf{significant (>4\%)}.

\begin{table*}[!t]
\caption{
Model performance on general tasks under weight-only quantization.
}
\vspace{-10pt}
\begin{center}
\resizebox{1\linewidth}{!}{
\begin{tabular}{c|c|c|cccccccc}
\toprule
\textbf{Model}    & \textbf{Setting}            & \textbf{Method} & \textbf{WinoGrande} & \textbf{PIQA} & \textbf{ARC-C} & \textbf{ARC-E} & \textbf{Hellaswag} & \textbf{MMLU 5-shot} & \textbf{Avg} & \textbf{Drop} \\ \midrule
\multirow{5}{*}{\textbf{LLaDA-8B}}                                                                          
& FP Model                    & -               & 69.9                & 74.6          & 46.4           & 71.1  & 70.7     & 65.7     & 65.5  & -    \\ \cmidrule{2-11} 
& \multirow{2}{*}{W4A16 g128} & GPTQ            & 69.7                & 73.9          & 47.9           & 72.5  & 70.4     & 64.7     & 65.3  & \cellcolor{green!10}$\downarrow$ 0.3\% \\
&                             & AWQ             & 67.3                & 70.3          & 44.5           & 73.4  & 68.4     & 65.6     & 63.2  & \cellcolor{orange!10}$\downarrow$ 3.5\% \\ \cmidrule{2-11} 
& \multirow{2}{*}{W3A16 g128} & GPTQ            & 67.2                & 73.3          & 45.7           & 71.1  & 68.8     & 63.5     & 63.7  & \cellcolor{orange!10}$\downarrow$ 2.7\% \\
&                             & AWQ             & 66.4                & 69.2          & 42.8           & 71.8  & 66.4     & 64.0     & 61.8  & \cellcolor{red!10}$\downarrow$ 5.6\% \\ \midrule
\multicolumn{1}{c|}{\multirow{5}{*}{\textbf{\begin{tabular}[c]{@{}c@{}}LLaDA-8B\\ -Instruct\end{tabular}}}} & FP Model                    & -               & 70.2                & 71.3          & 54.3           & 75.9  & 68.6               & 64.0                 & 65.7          & - \\ \cmidrule{2-11} 
\multicolumn{1}{c|}{}                                                                                       
& \multirow{2}{*}{W4A16 g128} & GPTQ            & 69.2                & 74.2          & 54.8           & 77.5  & 68.3     & 63.4     & 66.0  & \cellcolor{green!10}$\uparrow$ 0.3\%       \\
\multicolumn{1}{c|}{}                                                                                       
&                             & AWQ             & 68.8                & 71.0          & 53.3           & 76.1  & 68.1     & 63.4     & 64.9  & \cellcolor{green!10}$\downarrow$ 0.1\% \\ \cmidrule{2-11} 
\multicolumn{1}{c|}{}                                                                                       
& \multirow{2}{*}{W3A16 g128} & GPTQ            & 67.4                & 73.7          & 50.7           & 76.4  & 66.7     & 62.1     & 64.1  & \cellcolor{orange!10}$\downarrow$ 2.4\% \\
\multicolumn{1}{c|}{}                                                                                       
&                             & AWQ             & 66.3                & 69.8          & 50.5           & 74.7  & 66.4     & 62.3     & 63.1  & \cellcolor{orange!10}$\downarrow$ 4.0\% \\ \bottomrule
\end{tabular}
}
\end{center}
\label{tab:weight-only-general}
\end{table*}

\begin{table*}[!t]
\caption{
Model performance on general tasks under weight-activation quantization.
}
\vspace{-10pt}
\begin{center}
\resizebox{0.95\linewidth}{!}{
\begin{tabular}{c|c|c|ccccccc}
\toprule
\textbf{Model}                                                                         & \textbf{Setting}      & \textbf{Method} & \textbf{WinoGrande} & \textbf{PIQA} & \textbf{ARC-C} & \textbf{ARC-E} & \textbf{MMLU 5-shot} & \textbf{Avg} & \textbf{Drop} \\ \midrule
\multirow{7}{*}{\textbf{LLaDA-8B}}                                                     
& FP Model              & -               & 69.9                & 74.6          & 46.4           & 71.1           & 65.7          & 65.5         & -             \\ \cmidrule{2-10}
                                                                                       
& \multirow{3}{*}{W8A8} & SmoothQuant     & 67.7                & 70.8          & 45.5           & 70.5           & 65.0          & 63.9         & \cellcolor{orange!10}$\downarrow$ 2.5\%         \\
&                       & QuaRot          & 68.6                & 71.1          & 45.2           & 70.8           & 66.1          & 64.4         & \cellcolor{orange!10}$\downarrow$ 1.8\%         \\
&                       & DuQuant         & 67.9                & 70.4          & 45.9           & 71.4           & 66.0          & 64.3         & \cellcolor{orange!10}$\downarrow$ 1.9\%         \\ \cmidrule{2-10} 
& \multirow{3}{*}{W4A4} & SmoothQuant     & 49.4                & 58.8          & 29.2           & 40.9           & 27.1          & 41.1         & \cellcolor{red!10}$\downarrow$ 37.3\%        \\
&                       & QuaRot          & 63.4                & 68.1          & 43.7           & 69.2           & 61.8          & 59.2         & \cellcolor{red!10}$\downarrow$ 6.6\% \\
&                       & DuQuant         & 64.9                & 69.3          & 42.8           & 70.0           & 64.0          & 62.2         & \cellcolor{red!10}$\downarrow$ 5.1\%         \\ \midrule
\multirow{7}{*}{\textbf{\begin{tabular}[c]{@{}c@{}}LLaDA-8B\\ -Instruct\end{tabular}}} & FP Model              & -               & 70.2                & 71.3          & 54.3           & 75.9           & 64.0          & 67.1         & -             \\ \cmidrule{2-10} 
& \multirow{3}{*}{W8A8} & SmoothQuant     & 69.6                & 72.1          & 53.5           & 75.9           & 64.0          & 67.0         & \cellcolor{green!10}$\downarrow$ 0.2\%         \\
&                       & QuaRot          & 69.1                & 71.3          & 54.1           & 76.2           & 64.1          & 67.0         & \cellcolor{green!10}$\downarrow$ 0.3\%         \\
&                       & DuQuant         & 68.8                & 71.8          & 54.6           & 76.2           & 63.6          & 67.0         & \cellcolor{green!10}$\downarrow$ 0.2\%         \\ \cmidrule{2-10} 
& \multirow{3}{*}{W4A4} & SmoothQuant     & 52.3                & 65.1          & 34.0           & 54.0           & 32.8          & 47.7         & \cellcolor{red!10}$\downarrow$ 29.0\%        \\
&                       & QuaRot          & 65.2                & 69.8          & 51.3           & 75.1           & 61.1          & 64.5         & \cellcolor{orange!10}$\downarrow$ 3.9\%         \\
&                       & DuQuant         & 66.4                & 72.2          & 52.7           & 74.8           & 61.2          & 65.4         & \cellcolor{orange!10}$\downarrow$ 2.5\%         \\ \bottomrule
\end{tabular}
}
\end{center}
\label{tab:wa-general}
\end{table*}

\subsection{Ideal Quantization Bit Precision (RQ1)}

\paragraph{4-bit is the Recommended Choice for Weight-Only Quantization.}

We observe that both GPTQ and AWQ perform well on general commonsense QA and math tasks under 4-bit quantization (Table~\ref{tab:weight-only-general} and Table~\ref{tab:weight-only-math-code}). In most cases, the performance degradation remains within the \textit{negligible} to \textit{moderate} range (i.e., <4\%).
For example, 4-bit GPTQ-quantized LLaDA-8B-instruct slightly improves the average accuracy on six QA tasks from 65.7\% to 66.0\%, and shows only a minor drop of 0.6\% on the MATH and GSM8K benchmarks.
In contrast, reducing the quantization bit-width to 3-bit leads to a \textit{significant} performance drop, particularly on math and code generation tasks, as shown in Table~\ref{tab:weight-only-math-code}.
Therefore, we recommend 4-bit quantization as the standard configuration for weight-only quantization of diffusion-based LLMs. The development of more robust 3-bit quantization methods remains an open research direction.

\paragraph{Weight-Activation Quantization: 8-bit is Tolerable, While 4-bit Remains Challenging.}
As shown in Table~\ref{tab:wa-general} and Table~\ref{tab:wa-math-code}, quantizing LLaDA models to W8A8 results in only minor performance degradation, largely independent of the specific quantization method. This suggests that even simple techniques such as SmoothQuant are effective in mitigating activation outliers in LLaDA models, leading to nearly lossless quantized variants.
However, reducing precision to W4A4 introduces a sharp performance drop across most benchmarks. 
In the majority of cases, performance degradation exceeds the \textit{significant} threshold (>4\%). For instance, SmoothQuant experiences a drop of over 20\% across all evaluated tasks, indicating that the simple rebalancing between weights and activations is insufficient under low-precision settings for dLLMs.
The degradation is especially pronounced in base models, with accuracy drops exceeding 10\% on code generation tasks and math reasoning-heavy benchmarks.
These results highlight the difficulty of achieving effective 4-bit weight-activation quantization in dLLMs, and point to the need for more advanced techniques. Improving performance under this challenging setting remains an open research problem for the community.

\begin{table*}[!t]
\caption{
Model performance on mathematics and code tasks under weight-only quantization.
}
\vspace{-10pt}
\begin{center}
\resizebox{1\linewidth}{!}{
\begin{tabular}{c|l|c|cccc|cccc}
\toprule
\multicolumn{1}{c|}{\multirow{2}{*}{\textbf{Model}}}                                   & \multirow{2}{*}{\textbf{Setting}} & \multirow{2}{*}{\textbf{Method}} & \textbf{GSM8K (4-shot)} & \textbf{Math (0-shot)} & \multirow{2}{*}{\textbf{Avg}} & \multirow{2}{*}{\textbf{Drop}} & \textbf{HumanEval (0-shot)} & \textbf{MBPP (3-shot )} & \multirow{2}{*}{\textbf{Avg}} & \multirow{2}{*}{\textbf{Drop}} \\
\multicolumn{1}{l|}{}                                                                  &                                   &                                  & Gen Len 256             & Gen Len 256            &                   &            & Gen Len 512                 & Gen Len 512             &               &                \\ \midrule
\multirow{5}{*}{\textbf{LLaDA-8B}}                                                     
& FP Model                          & -             & 69.7        & 21.3       & 45.5     & -                                         & 32.9                        & 39.4                    & 36.2      & -                    \\ \cmidrule{2-11} 
& \multirow{2}{*}{W4A16 g128}       & GPTQ          & 68.5        & 21.3       & 44.9     & \cellcolor{orange!10}$\downarrow$ 1.4\%   & 28.7                        & 39.4                    & 34.0      & \cellcolor{red!10}$\downarrow$ 5.9\%    \\
&                                   & AWQ           & 67.4        & 20.6       & 44.0     & \cellcolor{orange!10}$\downarrow$ 3.2\%   & 29.9                        & 37.2                    & 33.5      & \cellcolor{red!10}$\downarrow$ 7.3\%    \\ \cmidrule{2-11}
& \multirow{2}{*}{W3A16 g128}       & GPTQ          & 63.3        & 13.4       & 38.4     & \cellcolor{red!10}$\downarrow$ 15.7\%   & 26.2                        & 35.4                    & 30.8      & \cellcolor{red!10}$\downarrow$ 14.8\%    \\
&                                   & AWQ           & 64.3        & 17.0       & 40.6     & \cellcolor{red!10}$\downarrow$ 10.7\%   & 28.1                        & 34.2                    & 31.1      & \cellcolor{red!10}$\downarrow$ 13.9\%     \\ \midrule
\multirow{5}{*}{\textbf{\begin{tabular}[c]{@{}c@{}}LLaDA-8B\\ -Instruct\end{tabular}}} & FP Model                          & -                                & 78.5                    & 33.5                   & 56.0        & -                   & 37.8                        & 37.4                    & 37.6                          \\ \cmidrule{2-11} 
& \multirow{2}{*}{W4A16 g128}       & GPTQ          & 78.8        & 32.4       & 55.6     & \cellcolor{green!10}$\downarrow$ 0.6\%   & 36.6                        & 33.8                    & 35.2      & \cellcolor{red!10}$\downarrow$ 6.4\%    \\
&                                   & AWQ           & 78.9        & 33.6       & 56.2     & \cellcolor{green!10}$\uparrow$ 0.5\%   & 37.1                        & 35.6                    & 36.4      & \cellcolor{orange!10}$\downarrow$ 3.2\%    \\ \cmidrule{2-11}
& \multirow{2}{*}{W3A16 g128}       & GPTQ          & 76.4        & 30.0       & 53.2     & \cellcolor{red!10}$\downarrow$ 5.0\%   & 34.2                        & 30.0                    & 32.1            & \cellcolor{red!10}$\downarrow$ 14.7\%    \\
&                                   & AWQ           & 76.3        & 30.1       & 53.2     & \cellcolor{red!10}$\downarrow$ 5.0\%   & 34.1                        & 31.8                    & 33.0            & \cellcolor{red!10}$\downarrow$ 12.4\%    \\ \bottomrule
\end{tabular}
}
\end{center}
\label{tab:weight-only-math-code}
\end{table*}

\begin{table*}[!t]
\caption{
Model performance on mathematics and code tasks under weight-activation quantization.
}
\vspace{-10pt}
\begin{center}
\resizebox{1.\linewidth}{!}{
\begin{tabular}{l|cc|cccc|cccc}
\toprule
\multirow{2}{*}{\textbf{Model}}                                                        
& \multirow{2}{*}{\textbf{Setting}} & \multirow{2}{*}{\textbf{Method}} & \textbf{GSM8K (4-shot)} & \textbf{Math (0-shot)} & \multirow{2}{*}{\textbf{Avg}} & \multirow{2}{*}{\textbf{Drop}} & \textbf{HumanEval (0-shot)} & \textbf{MBPP (3-shot )} & \multirow{2}{*}{\textbf{Avg}} & \multirow{2}{*}{\textbf{Drop}} \\
&                                   &                                  & Gen Len 256             & Gen Len 256            &                               &                                & Gen Len 512                 & Gen Len 512             &                               &                       \\ 
\midrule
\multirow{7}{*}{\textbf{LLaDA-8B}}                                                     
& FP Model                          & -                                & 69.7                    & 21.3                   & 45.5          & -                              & 32.9                        & 39.4                    & 36.2                          & -                     \\ \cmidrule{2-11} 
& \multirow{3}{*}{W8A8}             & SmoothQuant                      & 69.4                    & 20.2                   & 44.8          & \cellcolor{orange!10}$\downarrow$ 1.6\%                          & 27.4                        & 40.2                    & 33.8                       &  \cellcolor{red!10}$\downarrow$ 6.5\%                 \\
&                                   & QuaRot                           & 69.9                    & 20.7                   & 45.3          & \cellcolor{green!10}$\downarrow$ 0.4\%                      & 31.7                        & 40.6                    & 36.2                          &  \cellcolor{green!10}$\downarrow$ 0.0\%                 \\
&                                   & DuQuant                          & 70.7                    & 20.7                   & 45.7          & \cellcolor{green!10}$\uparrow$ 0.4\%                      & 33.5                        & 38.8                    & 36.2                          & \cellcolor{green!10}$\downarrow$ 0.0\%                 \\ \cmidrule{2-11} 
& \multirow{3}{*}{W4A4}             & SmoothQuant                      & 0.3                     & 2.0                    & 1.2           & \cellcolor{red!10}$\downarrow$ 97.4\%                         & 0.0                         & 0.0                     & 0.0                           & \cellcolor{red!10}$\downarrow$ 100.0\%               \\
&                                   & QuaRot                           & 62.9                    & 15.2                   & 39.1          & \cellcolor{red!10}$\downarrow$ 14.1\%                               & 23.8                        & 34.6                    & 29.2                          & \cellcolor{red!10}$\downarrow$ 19.3\%                \\
&                                   & DuQuant                          & 64.4                    & 14.8                   & 39.6          & \cellcolor{red!10}$\downarrow$ 13.0\%                         & 25.6                        & 33.6                    & 29.6                          & \cellcolor{red!10}$\downarrow$ 18.1\%                \\ \midrule
\multirow{7}{*}{\textbf{\begin{tabular}[c]{@{}l@{}}LLaDA-8B\\ -Instruct\end{tabular}}} 
& FP Model                          & -                                & 78.5                    & 33.5                   & 56.0          & -                              & 37.8                        & 37.4                    & 37.6                          & -                     \\ \cmidrule{2-11} 
& \multirow{3}{*}{W8A8}             & SmoothQuant                      & 78.2                    & 33.3                   & 55.7          & \cellcolor{green!10}$\downarrow$ 0.4\%                       & 37.2                        & 37.1                    & 38.6                          & \cellcolor{orange!10}$\downarrow$ 1.3\%                \\
&                                   & QuaRot                           & 78.9                    & 33.1                   & 56.0          & \cellcolor{green!10}$\uparrow$ 0.1\%                      & 35.4                        & 36.6                    & 36.0                          & \cellcolor{orange!10}$\downarrow$ 4.3\%                 \\
&                                   & DuQuant                          & 78.1                    & 33.3                   & 55.7          & \cellcolor{green!10}$\downarrow$ 0.5\%                       & 37.2                        & 37.4                    & 37.3                          & \cellcolor{green!10}$\downarrow$ 0.8\%                 \\ \cmidrule{2-11} 
& \multirow{3}{*}{W4A4}             & SmoothQuant                      & 2.7                     & 2.4                    & 2.6           & \cellcolor{red!10}$\downarrow$ 95.4\%                        & 0.0                         & 0.6                     & 0.3                           & \cellcolor{red!10}$\downarrow$ 99.2\%                \\
&                                   & QuaRot                           & 75.1                    & 29.9                   & 52.5          & \cellcolor{red!10}$\downarrow$ 6.2\%                        & 32.3                        & 32.8                    & 32.6                          & \cellcolor{red!10}$\downarrow$ 13.4\%                 \\
&                                   & DuQuant                          & 77.3                    & 30.7                   & 54.0          & \cellcolor{orange!10}$\downarrow$ 3.5\%                      & 34.8                        & 29.2                    & 32.0                       & \cellcolor{red!10}$\downarrow$ 14.9\%                \\ 
\bottomrule
\end{tabular}
}
\end{center}
\label{tab:wa-math-code}
\end{table*}

\subsection{Optimal Quantization Methods (RQ2)}

\paragraph{GPTQ Outperforms AWQ on Most Tasks}

As shown in Table~\ref{tab:weight-only-general}, GPTQ outperforms AWQ on average accuracy under both 3-bit and 4-bit quantization for LLaDA-8B and LLaDA-8B-instruct. This demonstrates the reliability and competitiveness of GPTQ, particularly on QA tasks.
This trend also holds for math reasoning tasks, except for the 3-bit quantization setting on LLaDA-8B, where both GPTQ and AWQ suffer critical performance degradation (>10\%).
We hypothesize that the suboptimal performance of AWQ may stem from the fact that activation outliers in the LLaDA model series are less prominent than in traditional LLMs. 
Since AWQ identifies the top 1\% of salient weights using activation-driven statistics, its effectiveness can be reduced when the outlier structure is weak in LLaDA models, thereby diminishing its advantage.
For code generation tasks, the situation becomes more complex. Both GPTQ and AWQ fail to maintain acceptable performance on the HumanEval and MBPP benchmarks under low-bit quantization. A more detailed analysis of these results is provided in Section~\ref{subsec:task}.
Notably, AWQ performs relatively better than GPTQ in the 3-bit configuration for code tasks, suggesting some resilience under extreme compression.
Considering all evaluations across task types and bit-widths, we recommend GPTQ as the safer and more generally effective choice for weight-only quantization of diffusion-based language models.

\paragraph{Rotation-Based Methods Achieve Leading Performance Under Weight-Activation Quantization.}

For both LLaDA-8B and LLaDA-8B-instruct, rotation-based methods—QuaRot and DuQuant—consistently outperform SmoothQuant across all evaluation tasks and quantization settings.
The advantage becomes especially pronounced under 4-bit weight-activation quantization, where SmoothQuant suffers a near-complete performance collapse on code and math tasks. In contrast, rotation-based approaches retain a non-trivial portion of model capability, highlighting their robustness in low-precision settings.
These results suggest that rotation transformations are more effective in mitigating activation outliers in dLLMs, which aligns with findings from prior studies in the LLM community~\citep{lin2024duquant,ashkboos2024quarot}.
When comparing QuaRot and DuQuant in detail, our experiments show that DuQuant consistently outperforms QuaRot across most scenarios.
For instance, on commonsense QA tasks, DuQuant achieves lower performance drops than QuaRot for both LLaDA-8B (5.1\% vs. 6.6\%) and LLaDA-8B-instruct (2.5\% vs. 3.9\%).
This observation remains consistent across math and code generation tasks.
Consequently, we recommend DuQuant as the most effective method for weight-activation quantization in diffusion-based language models.

\begin{table*}[!t]
\caption{
Evaluation of weight-only quantized Dream-7B on general tasks.
}
\vspace{-10pt}
\begin{center}
\resizebox{0.95\linewidth}{!}{
\begin{tabular}{c|c|c|cccccc}
\toprule
\textbf{Model}                     & \textbf{Setting}            & \textbf{Method} & \textbf{WinoGrande} & \textbf{PIQA} & \textbf{ARC-C} & \textbf{ARC-E} & \textbf{Avg} & \textbf{Drop} \\ \midrule
\multirow{5}{*}{\textbf{Dream-7B}} & FP Model                    & -               & 68.4                & 74.4          & 59.0           & 83.1           & 71.2         & -             \\ \cmidrule{2-9} 
& \multirow{2}{*}{W4A16 g128} & GPTQ            & 68.2                & 73.9          & 58.1           & 82.1           & 70.6         & \cellcolor{green!10}$\downarrow$ 0.8\%         \\
&                             & AWQ             & 65.2                & 69.6          & 55.8           & 82.0           & 68.2         & \cellcolor{red!10}$\downarrow$ 4.3\%         \\ \cmidrule{2-9} 
& \multirow{2}{*}{W3A16 g128} & GPTQ            & 63.3                & 69.6          & 49.9           & 73.4           & 64.1         & \cellcolor{red!10}$\downarrow$ 10.1\%        \\
&                             & AWQ             & 62.8                & 67.7          & 50.6           & 74.5           & 63.9         & \cellcolor{red!10}$\downarrow$ 10.3\%        \\ \bottomrule
\end{tabular}
}
\end{center}
\label{tab:dream}
\end{table*}


\subsection{Influence of Task Categories on Quantization (RQ3)}
\label{subsec:task}

\paragraph{Quantization is More Challenging for Math and Code Tasks.}

Compared to general-purpose benchmarks—primarily QA tasks as shown in Table~\ref{tab:weight-only-general} and Table~\ref{tab:wa-general}—quantized models experience significantly larger performance drops on math and code tasks, illustrated in Table~\ref{tab:weight-only-math-code}
and Table~\ref{tab:wa-math-code}.

For \textbf{math reasoning} tasks, both AWQ and GPTQ exhibit substantial degradation under 3-bit quantization (see Table~\ref{tab:weight-only-math-code}), despite maintaining competitive performance on general QA benchmarks. A similar trend is observed for rotation-based methods under W4A4 configurations.
This degradation may be attributed to the multi-step reasoning nature of math problems, which amplifies the cumulative effect of quantization errors. In such tasks, precise intermediate representations are critical; even small perturbations introduced by low-bit quantization can propagate and compound, ultimately leading to incorrect final answers.

In \textbf{code generation} tasks, the challenges are even more pronounced. Under 4-bit quantization, GPTQ and AWQ show performance drops exceeding 5\%, while QuaRot and DuQuant degrade by over 10\% under W4A4 for both LLaDA-8B and LLaDA-8B-instruct models. 
We also observe that the standard deviation on the HumanEval benchmark is relatively high, approximately 3\%, indicating that more robust and stable benchmarks may be needed to accurately assess code generation capabilities under quantization.
Code generation tasks often require the model to maintain long-range context and generate syntactically correct, semantically meaningful sequences. These demands are highly sensitive to the precision of both weights and activations. Quantization-induced distortion in attention patterns or token representations can disrupt code syntax or logic, causing severe performance degradation.

These observations highlight that math and code tasks impose stricter precision requirements than simpler retrieval-based or classification-style QA tasks. Maintaining accurate intermediate states, multi-hop logic, and long-context dependency are especially vulnerable under aggressive quantization.
Consequently, task-specific quantization strategies or adaptive precision control mechanisms may be necessary to improve the robustness of dLLMs on math and code benchmarks. This represents a critical direction for future research in efficient diffusion-based LLM deployment.

\subsection{Impact of Model Types (RQ4)}

\paragraph{Instruct-Tuned Models are More Robust than Base Models.}

We observe an interesting phenomenon: LLaDA-8B-instruct consistently exhibits smaller performance degradation than its base counterpart (LLaDA-8B) under nearly all quantization settings.
For instance, under general tasks, both DuQuant and QuaRot result in only minor accuracy drops for the instruct model, whereas the drop exceeds 5\% for the base model.
This trend remains consistent across more challenging math and code tasks.
For example, 3-bit quantized GPTQ and AWQ lead to performance degradation of approximately 5\% for the instruct variant, while the base model suffers drops as high as 10\%.

\textbf{Our Observations Hold Consistently across Different dLLMs.}
To assess the generality of our findings, we further evaluate various quantization methods on a different diffusion-based language model: Dream-7B.
As shown in Table~\ref{tab:dream}, both GPTQ and AWQ perform competitively under 4-bit quantization, while performance drops become more pronounced in the 3-bit setting. 
This observation reinforces our recommendation that 4-bit quantization offers a near-lossless trade-off between efficiency and performance.
Moreover, GPTQ consistently outperforms AWQ across nearly all benchmarks, suggesting that GPTQ is a more reliable choice across different types of dLLMs.
Notably, the 3-bit quantized models exhibit risk-level degradation even on general tasks, indicating that aggressive quantization may be more challenging for the Dream model series compared to LLaDA.
Due to resource constraints, we did not evaluate weight-activation quantization for Dream-7B. We leave this for future work as part of our ongoing exploration.

\section{Limitation and Future Work}

In this work, 
our primary focus is on evaluating downstream task performance of quantized dLLMs. 
Quantization offers an effective way to reduce memory consumption and accelerate inference.
However, fully integrating low-bit inference for diffusion LLMs remains challenging. Specifically, adapting existing LLM-optimized kernels to the architectural characteristics of diffusion LLMs involves substantial engineering effort, which we leave for future work.

We plan to continue this line of research along the following directions:
1). Expanded Evaluation: We will provide a more comprehensive evaluation across a broader set of dLLMs, tasks, and model sizes.
2). Stepwise Analysis: We aim to explore how the number of generation steps in diffusion decoding interacts with quantization levels, and whether step-aware quantization strategies can be beneficial.
3). Remasking Strategies: We intend to evaluate different remasking strategies under quantized settings, and provide practical guidance on selecting suitable quantization configurations.

We hope our work initiates further discussion and exploration in the community. To facilitate future research, we will release our code and implementation details to support the development and deployment of quantized diffusion LLMs.
\section{Conclusion}

This work provides the first in-depth investigation into the challenges and opportunities of applying post-training quantization (PTQ) to diffusion-based language models (dLLMs).
Through extensive empirical evaluation, we uncover several key findings: (1) activation outliers are prevalent across dLLMs and are critical barriers to low-bit quantization; (2) certain PTQ methods, GPTQ and DuQuant, demonstrate notable advantages under constrained settings; and (3) quantization behavior varies across tasks and model types, with instruct-tuned models showing greater resilience.
These findings offer practical guidance for designing more effective and robust quantization strategies.
Looking forward, we believe that our study lays the groundwork for future research in compression of dLLMs, enabling their deployment in real-world, resource-constrained environments.

\newpage
\clearpage

\appendix

\renewcommand\thefigure{\Alph{section}\arabic{figure}}
\renewcommand\thetable{\Alph{section}\arabic{table}}
\setcounter{figure}{0}
\setcounter{table}{0}

\section*{Appendix}

\section{Additional Implementation Details}
\label{appx:imple}
\paragraph{Weight-only Quantization Methods.} 
For GPTQ, we use 128 calibration samples from the WikiText-2 dataset with a sequence length of 2048. We adopt asymmetric quantization and set the group size to 128. The method is implemented using the AutoGPTQ repository\footnote{https://github.com/AutoGPTQ/AutoGPTQ.}.
For AWQ, we use 128 samples from the PileVal dataset with a sequence length of 512. 
We implement AWQ with the llm-awq repository\footnote{https://github.com/mit-han-lab/llm-awq.} and
apply the same settings as GPTQ, using asymmetric quantization and a group size of 128.

\paragraph{Weight-activation Quantization Methods.}
For SmoothQuant, we set the hyperparameter $\alpha = 0.5$ in the scaling equation to compute the diagonal matrix:
$s_j=\max({|\mathbf{X}_j|})^{\alpha}/\max({|\mathbf{W}_j|})^{1-\alpha}$. 
We use 128 calibration samples from the WikiText-2 dataset with a sequence length of 2048. Asymmetric per-tensor quantization is applied to weights, and per-channel quantization is applied to activations.
For QuaRot, we follow the original configuration by preserving 16-bit precision for query states, and applying symmetric activation quantization. We also use WikiText-2 (128 samples, sequence length 2048) as the calibration dataset.
For DuQuant, we use the same calibration setup and $\alpha$ value as SmoothQuant. Additionally, we apply activation and weight clipping ratios of 0.9 and 0.8, respectively. The rotation step is set to 256, and the block size is 128.
\paragraph{General Tasks.}
We employ the lm-evaluation-harness\footnote{https://github.com/EleutherAI/lm-evaluation-harness}
 repository to benchmark models across all tasks. For general tasks other than MMLU, we adopt a 0-shot setting with 128 Monte Carlo samples. For MMLU, we use a 5-shot setting with a single Monte Carlo sample. To evaluate LLaDA models, we configure the diffusion steps, block size, and generation length to 1024, set the classifier-free guidance (CFG) scale and temperature to 0.0, and apply the \emph{low confidence} remasking strategy. For Dream, we set the maximum number of new tokens to 128, the CFG scale to 1.0, the temperature to 0.0, and the top-\emph{p} threshold (probability of retaining generated tokens) to 0.95.

\paragraph{Mathematics and Code Tasks.}
The benchmarking details of LLaDA on mathematics and code tasks are provided in Tab.~\ref{tab:tasks_details}. All other configurations remain the same as in the general tasks.

\begin{table*}[h]
\caption{Configuration for mathematics and code tasks.}
\label{tab:tasks_details}
\vspace{-10pt}
\begin{center}
\resizebox{0.7\linewidth}{!}{
\begin{tabular}{c|cccc}
\toprule
\textbf{Dataset}    & \# fewshots & generation length & diffusion steps & block size \\\midrule
GSM8K & 4 & 256 & 256&32\\
Math & 0 & 256 & 256&64\\
HumanEval & 0 & 512 & 512&32\\
MBPP &3& 512 & 512&32\\
\bottomrule
\end{tabular}
}
\vspace{-0.5cm}
\end{center}
\end{table*}

\section{More Visualizations}

 \begin{figure*}[h]
    \centering
    \includegraphics[width=1.\linewidth]{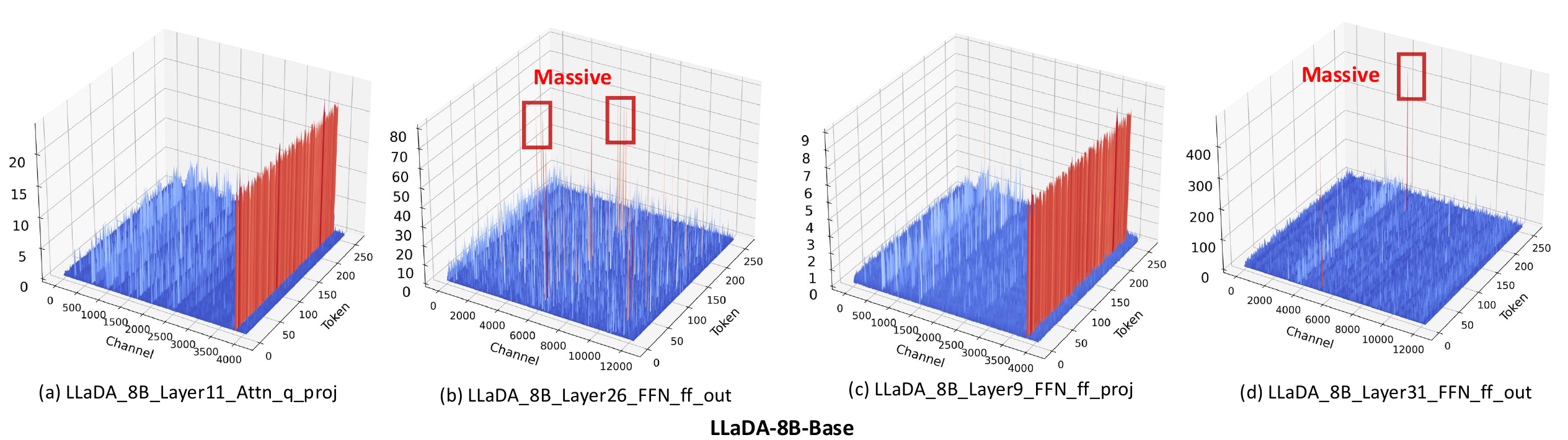
    }
    \caption{
More visualizations of activation outliers in LLaDA-8B-Base.
    }
    \label{fig:appx_llada}    
\end{figure*}

 \begin{figure*}[h]
    \centering
    \includegraphics[width=1.\linewidth]{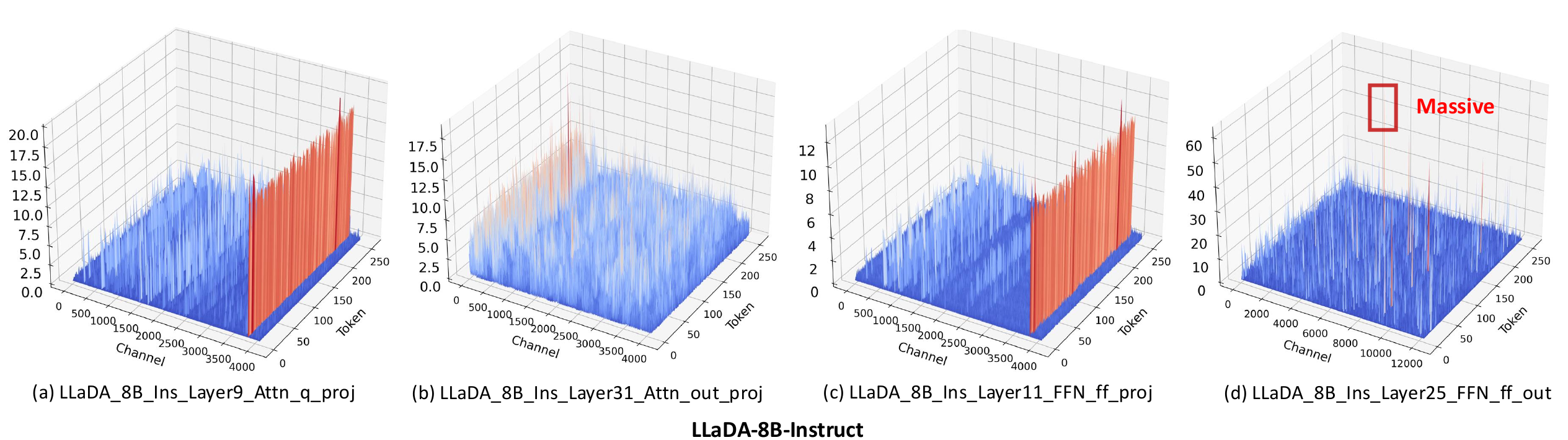
    }
    \caption{
More visualizations of activation outliers in LLaDA-8B-Instruct.
    }
    \label{fig:appx_llada_ins}    
\end{figure*}

\newpage

\bibliography{nips}
\bibliographystyle{plainnat}

\end{document}